\documentclass[runningheads]{llncs}
\usepackage[T1]{fontenc}
\usepackage{graphicx}
\usepackage{amsmath}
\usepackage{float}
\usepackage[numbers,sort&compress]{natbib}
\usepackage{hyperref}
\hypersetup{
colorlinks   = true,
citecolor    = blue,
urlcolor    =  blue
}
\usepackage{color}

\begin{document}
\title{DESS: DeBERTa Enhanced Syntactic-Semantic Aspect Sentiment Triplet Extraction}
%
%
\author{Vishal Thenuwara\orcidID{0009-0005-9932-0454} \and
Nisansa de Silva\orcidID{0000-0002-5361-4810}}

\authorrunning{V. Thenuwara and N. de Silva}

\institute{Department of Computer Science \& Engineering,\\
University of Moratuwa, Sri Lanka\\
\email{\{vishal.23,NisansaDdS\}@cse.mrt.ac.lk}\\
}
 \maketitle              
%

\begin{abstract}
Fine-grained sentiment analysis faces ongoing challenges in Aspect Sentiment Triple Extraction (ASTE), particularly in accurately capturing the relationships between aspects, opinions, and sentiment polarities. While researchers have made progress using BERT and Graph Neural Networks, the full potential of advanced language models in understanding complex language patterns remains unexplored. We introduce DESS, a new approach that builds upon previous work by integrating DeBERTa's enhanced attention mechanism to better understand context and relationships in text. Our framework maintains a dual-channel structure, where DeBERTa works alongside an LSTM channel to process both meaning and grammatical patterns in text. We have carefully refined how these components work together, paying special attention to how different types of language information interact. When we tested DESS on standard datasets, it showed meaningful improvements over current methods, with F1-score increases of 4.85, 8.36, and 2.42 in identifying aspect opinion pairs and determining sentiment accurately. Looking deeper into the results, we found that DeBERTa's sophisticated attention system helps DESS handle complicated sentence structures better, especially when important words are far apart. Our findings suggest that upgrading to more advanced language models when thoughtfully integrated, can lead to real improvements in how well we can analyze sentiments in text. The implementation of our approach is publicly available at:~\url{https://github.com/VishalRepos/DESS}.

\end{abstract}
\section{Introduction}
Aspect Sentiment Triple Extraction (ASTE) has emerged as a crucial task in fine-grained sentiment analysis~\cite{de2025linguistic,gunathilaka2025automatic}, aiming to identify opinion targets, their associated expressions, and corresponding sentiment polarities from unstructured text~\cite{jayakody2024aspect,jayakody2024enhanced,jayakody2024instruct}. For instance, in the review "\textit{The camera quality is excellent but battery life disappoints,}" ASTE needs to extract three key elements: \textit{aspects} (camera quality, battery life), \textit{opinions} (excellent, disappoints), and their associated \textit{sentiments} (positive, negative).
Recent approaches have primarily relied on sequence labeling and span-based methods to tackle this challenge. While sequence labeling techniques effectively identify aspect and opinion terms, they often struggle with complex relationships between multiple aspects and opinions within the same sentence~\cite{jayakody2024aspect}. Span-based methods improved upon this by directly modeling the relationships between text spans, but their effectiveness remains limited by their inability to fully capture the rich syntactic and semantic information inherent in natural language.~\cite{xu2021learning}
Several researchers have attempted to address these limitations by incorporating Graph Neural Networks (GNNs) to model syntactic dependencies~\cite{chen2022enhanced}. However, these methods typically focus on either syntactic or semantic information, missing the opportunity to leverage their complementary nature. Additionally, existing approaches often rely on BERT-based encoders, which, while powerful, have known limitations in handling disentangled attention patterns and relative position modeling~\cite{zhu2022deep,wu2020grid}.A key motivation of our research effort is to replace the conventional BERT encoder with the more efficient DeBERTa encoder to overcome these architectural limitations. 
We propose DESS (DeBERTa Enhanced Syntactic-Semantic), a novel framework that addresses these challenges by leveraging DeBERTa's enhanced attention mechanism alongside carefully designed syntactic and semantic processing components. Our approach differs from previous work in three key aspects. First, we upgrade the backbone encoder to DeBERTa-v3, taking advantage of its disentangled attention patterns for better context understanding. Second, we maintain separate but interconnected channels for processing syntactic and semantic information, allowing each type of information to develop independently before integration. Third, we introduce a refined heterogeneous feature interaction module that dynamically balances the contribution of syntactic and semantic features.
Through extensive experimentation on benchmark datasets, we demonstrate that DESS not only achieves state-of-the-art performance but also shows particular strength in handling complex cases where aspects and opinions are separated by long distances or connected through intricate syntactic structures.

\section{Related Work}

Aspect-Based Sentiment Analysis (ABSA) is a critical area in Natural Language Processing (NLP), focusing on extracting fine-grained sentiment information. Within this domain, Aspect Sentiment Triplet Extraction (ASTE) aims to identify aspect terms, opinion terms, and their associated sentiment polarity from text. Early approaches to ASTE often utilized pipeline methods, which, while effective, suffered from error propagation issues \cite{peng2020knowing}. To mitigate these challenges, end-to-end models were developed, enhancing the extraction process by jointly learning the components of the triplets \cite{xu2020position}. 

\subsection{Aspect Sentiment Triplet Extraction (ASTE)}
The ASTE task was first introduced by \citet{peng2020knowing}, proposing a pipeline-based extraction method that performed aspect and opinion term extraction separately, followed by sentiment classification. However, the reliance on a multi-stage process led to cumulative errors. To address these limitations, \citet{wu2020grid} proposed the Grid Tagging Scheme (GTS), reformulating ASTE as a structured prediction task. Similarly, \citet{li2022span} introduced the Span-Shared Joint Extraction (SSJE) model, improving performance but struggling with multi-aspect sentence structures and implicit aspects.

\subsection{Graph Neural Networks (GNN) for ASTE}
Recent advancements have integrated Graph Neural Networks (GNNs) into ASTE models to better capture syntactic and semantic dependencies. \citet{chen2022enhanced} proposed an Enhanced Multi-Channel Graph Convolutional Network (EMC-GCN) that leverages syntactic features to improve sentiment analysis, demonstrating the effectiveness of incorporating structural linguistic information. Similarly, \citet{peng2023ptgcn} introduced a Prompt-Based Tri-Channel Graph Convolution Neural Network (PT-GCN), which employs prompt-based attention mechanisms to refine extraction by emphasizing relevant syntactic dependencies. While these models have shown promise in capturing complex relationships within text, challenges remain in modeling long-range dependencies and handling implicit aspects.
\subsection{Transformer-Based Approaches}
Transformer-based models have also played a crucial role in advancing ASTE. \citet{naglik2024aste} presented the ASTE-Transformer, modeling dependencies in aspect-sentiment triplet extraction and highlighting the importance of intricate relationships within text. However, traditional transformers struggle with encoding positional information effectively, leading to errors in aspect-opinion linking. To address these shortcomings, \citet{hedeberta} introduced DeBERTa (Decoding-enhanced BERT with Disentangled Attention), which separates content and positional embeddings to improve dependency modelling. The disentangled attention mechanism allows for a more refined understanding of syntactic and semantic relationships, making it particularly suitable for ASTE.

\section{Methodology}

\begin{figure*}[!htb]
    \centering
    \includegraphics[width=\textwidth]{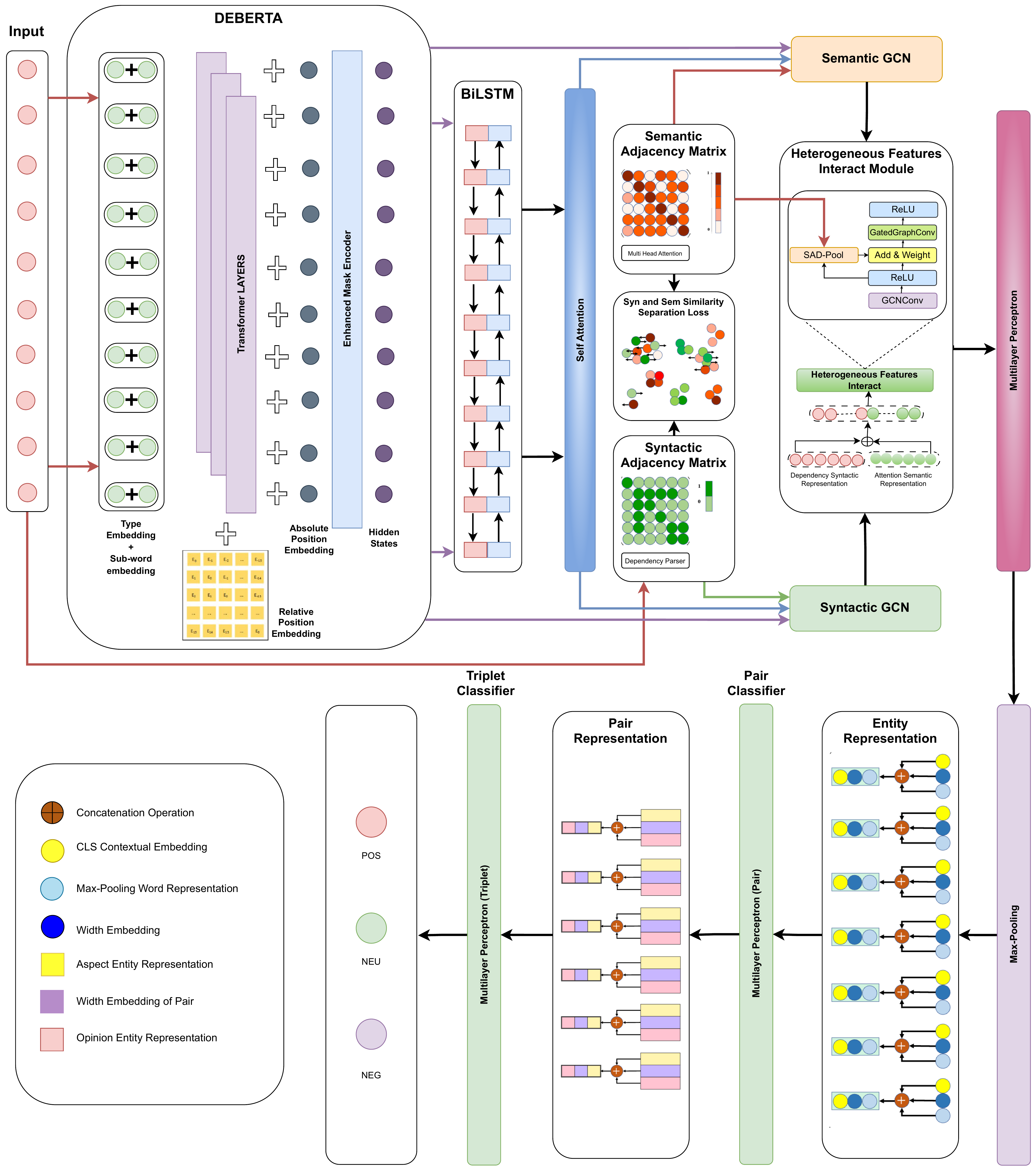}
    \caption{DESS Architecture}
    \label{fig:DESS-architecture}
\end{figure*}

In this work, we propose an improved DeBERTa-based Dual Encoder model (DESS) that advances Aspect Sentiment Triplet Extraction (ASTE) beyond traditional BERT-based approaches as shown in Figure~\ref{fig:DESS-architecture}. By replacing BERT with DeBERTa in the D2E2S framework~\cite{zhao2024dual}, our model benefits from disentangled attention and enhanced positional encodings, leading to improved contextual understanding and linguistic sensitivity. This advancement supports three key capabilities: (1) detecting implicit aspects in complex sentences, (2) reinforcing semantic links between aspects and opinions, and (3) more accurate sentiment polarity classification under context-specific conditions.

To further strengthen the architecture, we integrate Graph Neural Networks (GNNs) and a Heterogeneous Feature Interaction Module (HFIM), enabling the model to capture both syntactic dependencies and semantic relationships. This results in a comprehensive structure capable of extracting complete opinion triplets with improved accuracy across diverse datasets.

\subsection{Task Definition}

The ASTE task aims to extract opinion triplets $(a, o, s)$, where $a$ denotes the aspect term, $o$ is the opinion term, and $s$ indicates the sentiment (\textit{positive}, \textit{neutral}, or \textit{negative}). For example, in the sentence \textit{"The price is reasonable, but the service is poor"}, the triplets are: \textit{(price, reasonable, positive)} and \textit{(service, poor, negative)}.

This task is challenging due to the need for modeling both syntactic and semantic dependencies within a sentence. The original D2E2S model addressed this using a dual-encoder architecture—semantic and syntactic channels—which we enhance through DeBERTa integration.

\subsection{Model Integration and Adjustments}

We evaluate the effect of replacing BERT in D2E2S with three DeBERTa variants: \texttt{DeBERTa V3-Base} (86M parameters), \texttt{DeBERTa V3-Large} (304M parameters), and \texttt{DeBERTa V2-xxLarge} (1.5B parameters). Each variant replaces BERT in both the semantic and syntactic channels. Several adjustments are made to ensure compatibility: the BERT tokenizer is replaced with the DeBERTa tokenizer to align with DeBERTa's disentangled attention; hidden state sizes are updated to match each DeBERTa variant (768, 1024, and 1536 respectively); and the architecture is adapted to handle the increased representation capacity.

We also tuned hyperparameters—learning rates, batch sizes, warmup ratios—for each model variant. Smaller batch sizes and lower learning rates are used for larger models to manage memory and avoid overfitting. Table~\ref{tab:model-parameters} shows the training parameters.

\begin{table}[h]
\centering
\caption{Training Parameters for DeBERTa Variants}
\label{tab:model-parameters}
\begin{tabular}{lccc}
\hline
\textbf{Model} & \textbf{Learning Rate} & \textbf{Batch Size} & \textbf{Warmup Ratio} \\
\hline
V3-Base & $5 \times 10^{-5}$ & 16 & N/A \\
V3-Large & $5 \times 10^{-6}$ & 12 & 0.2 \\
V2-xxLarge & $5 \times 10^{-6}$ & 8 & 0.2 \\
\hline
\end{tabular}
\end{table}

\subsection{Dataset Splits and Training Protocol}
\label{sec:dataset-splits}

For each of the four ASTE-Data-V2 subsets (Restaurant-14, Laptop-14, Restaurant-15, Restaurant-16), we follow the conventional three-way data partitioning into training, development, and test splits. During the model fitting stage, only the training set is exposed to the optimizer: all gradient updates for the DeBERTa encoders, GNN layers, HFIM, and associated parameters are computed exclusively on these examples. After each epoch of training, we evaluate the current model checkpoint on the development split to compute validation metrics—primarily F$_1$ and loss—which serve two purposes: guiding early stopping when performance ceases to improve, and informing hyperparameter decisions such as learning rate, batch size, and warmup ratio. Crucially, no back-propagation or parameter updates are performed on the development examples; they remain strictly held out for unbiased validation. Only after selecting the best-performing checkpoint based on development-set F$_1$ do we perform a single, final evaluation on the test split to report our Precision, Recall, and F$_1$ results. This protocol ensures that neither development nor test data influence the learned parameters, preserving the integrity of our experimental comparisons.

\section{Experiments}
\subsection{Settings}

Our experiments use standard SemEval benchmark datasets: \textit{14LAP} (laptop reviews) and \textit{14RES}, \textit{15RES}, \textit{16RES} (restaurant reviews). We implemented DESS using PyTorch, with \texttt{DeBERTa-v3-base} as the main encoder, utilizing 768-dimensional hidden representations. Training employed the \texttt{AdamW} optimizer with a learning rate of $2e-5$ for the encoder and $1e-4$ for other components, decayed via a linear scheduler with a $0.1$ warmup ratio.
The model architecture includes a bidirectional LSTM with 2 layers of 384 hidden units and a 0.5 dropout rate. Both syntactic and semantic GCNs operate on 384-dimensional hidden states with a 0.3 dropout. Ps are limited to 128 tokens, with batch size set to 8. Size embeddings use 25 dimensions, and entity pairs are capped at 100 for memory efficiency.
Training runs for 120 epochs, with performance monitored on validation data. The loss function combines cross-entropy with a KL divergence term to align syntactic and semantic features. Experiments were conducted on a single NVIDIA A100 GPU (40GB).
Performance evaluation follows standard exact-match criteria for entity boundaries and sentiment labels, reporting precision, recall, and F1 scores for aspect-opinion extraction and sentiment classification.

\subsection{Datasets}
The experiments in this research utilize the D2 dataset, refined by~\citet{xu2020position}, addressing missing triplets and conflicting sentiment labels. D2 consists of four benchmark datasets: three on restaurant reviews (14RES, 15RES, 16RES) and one on laptop reviews (14LAP), originally derived by~\citet{peng2020knowing}.
Table~\ref{tab:datasets} in Appendix~\ref{app:stats} presents dataset statistics. 
It should be noted that sentiment analysis reveals a dominance of positive sentiment, with 14RES training data containing 1,692 positive instances versus 480 negative and 166 neutral cases. This trend persists across all datasets, reflecting a bias toward positive expressions in customer reviews.
The structured dataset design ensures balanced training, development, and test sets, providing a robust foundation for evaluating model performance across domains and sentiment variations, offering meaningful real-world insights.

\subsection{Main Results}

\begin{table*}[!htb]
	\caption{Precision (\%), Recall (\%) and F1 score (\%) on the benchmark D2 (\citet{xu2020position}). All baseline results from the original papers. Models with '†' are results reproduced by us.}
	\label{merged_result}
	\footnotesize
	\centering
	\resizebox{\textwidth}{!}{\begin{tabular}{l|ccc|ccc|ccc|ccc}
			\hline
                \multicolumn{1}{c|}{Model}
								& \multicolumn{3}{c|}{14LAP}                       & \multicolumn{3}{c|}{14RES}               & \multicolumn{3}{c|}{15RES}                       & \multicolumn{3}{c}{16RES}                        \\ \cline{2-13} 
								& P              	& R              	& F1             	& P      	& R              	& F1             & P              & R              & F1             & P              & R              & F1             \\ \hline

			CMLA+~\cite{wang2017multi}            	    & 30.10          	& 36.90          & 33.20          	& 39.20  & 39.80          & 37.00          & 41.30          & 42.10          & 41.70          & 41.3           &42.1            & 41.7               \\
			RINANTE+ ~\cite{peng2020knowing}       		& 21.70          	& 18.70          & 20.10          	& 29.90  & 30.10          & 30.00          & 25.70          & 22.30          & 23.90          & 25.7           & 22.3           & 23.9               \\
                WhatHowWhy~\cite{peng2020knowing}              & 37.38             & 50.38          & 42.87            & 43.24  & 48.07          & 63.66          & 51.46          & 57.51          & 52.32          & 46.96          & 64.24          & 54.21          \\
                TOP                     & 57.84             & 59.33          & 58.58            & 54.53  & 63.59          & 73.44          & 68.16          & 63.30          & 58.59          & 63.57          & 71.98          & 67.52          \\
			Li-unified-R~\cite{peng2020knowing}   			& 40.60          	& 43.40          & 42.30          	& 44.70  & 51.40          & 47.80          & 37.30          & 54.50          & 44.30          & 40.60          & 44.30          & 42.30          \\
			Peng-two-stage~\cite{mukherjee2023contraste}  		& 37.40          	& 50.40          & 43.90          	& 48.10  & 57.50          & 52.30          & 47.00          & 64.20          & 54.20          & 37.40          & 50.40          & 43.90          \\            
                ChatGPT         		& 35.10          	& 48.90          & 41.80          	& 41.90  & 60.60          & 49.50          & 44.90          & 61.70          & 52.00          & 35.10          & 48.90          & 41.80          \\
			JET-BERT~\cite{zhu2022deep}        		& 55.40          	& 47.30          & 51.30         	& 64.50  & 52.00          & 57.00          & 70.40          & 58.40          & 64.80          & 55.40          & 47.30          & 51.30          \\
			PASTE ~\cite{mukherjee2021paste}          		& 55.00          	& 51.60          & 53.80          	& 58.30  & 56.70          & 57.50          & 65.50          & 64.40          & 64.90          & 55.00          & 51.60          & 53.80          \\
			JETt~\cite{zhu2022deep}           			& 51.48             & 42.65          & 46.65            & 70.20  & 53.02          & 60.41          & 62.14          & 47.25          & 53.68          & 72.12          & 57.20          & 63.41          \\
			JETo~\cite{zhu2022deep}           			& 58.47             & 43.67          & 50.00            & 67.97  & 60.32          & 63.92          & 58.35          & 51.43          & 54.67          & 64.77          & 61.29          & 62.98          \\
						
                Unified-BART-ABSA 	    & 65.40        	    & 65.00          & 65.20          	& 59.10  & 59.40          & 59.30          & 66.60          & 68.70          & 67.70          & 65.40          & 65.00          & 65.20          \\
			BMRC~\cite{liu2022robustly}            		& 70.60          	& 61.80          & 66.20          	& 68.50  & 53.40          & 60.10          & 71.20          & 61.10          & 66.10          & 70.60          & 61.80          & 66.20          \\
                PARAPHRASE      		& 71.10          	& 71.90          & 71.50          	& 60.40  & 63.90          & 62.70          & 70.10          & 73.90          & 72.50          & 71.10          & 71.90          & 71.50          \\
			SBSK-ASTE~\cite{feng2023improving}       		& 74.60          	& 71.50          & 73.10          	& 65.30  & 63.70          & 64.50          & 70.80          & 72.00          & 71.40          & 74.60          & 71.50          & 73.10          \\
			Span-BiDir~\cite{mukherjee2023contraste}      		&\textbf{76.40}     & 72.40          &74.30 		    & 69.90  & 60.40          & 64.80 	       & 71.60          & 72.60          & 72.10          & 76.40          & 72.40          & 74.30           \\
                Unified                 & 61.41            	&56.19          &58.69  		    & 65.52  & 64.99          & 65.25          & 59.14          & 59.38          & 59.26          & 66.60          & 68.68          & 67.62             \\
                BARTABSA                & 61.41           	& 56.19          &58.69           	& 65.52  & 64.99           &65.25           &59.14           &59.38           &59.26           &66.6            &68.68           &67.62     \\             
			GAS             		& 65.00          	& 69.50          & 67.20          	& 56.10  & 61.80          & 58.80          & 66.10          & 68.70          & 67.40          & 65.00          & 69.50          & 67.20          \\
                GTS-BERT~\cite{wu2020grid}                & 59.40           	& 51.94          & 55.42           	& 68.09  & 69.54          & 68.81          & 59.28          & 57.93          & 58.60          & 68.32          & 66.86          & 67.58          \\
			Double-Encoder~\cite{wang2021explicit}            & 62.12             & 56.38          & 59.11            & 67.95  & 71.23          & 69.55          & 58.55          & 60.00          & 59.27          & 70.65          & 70.23          & 70.44              \\
                TGA+SFI~\cite{wang2021explicit}                 & 65.25             & 53.79          & 58.98             & 71.75 & 70.52          & 71.13          & 62.77          & 59.79          & 61.25          & 68.20           & 69.26         & 68.73             \\
                EMC-GCN~\cite{chen2022enhanced}       		    & 61.70          	& 56.30          & 58.80          	& 62.50  & 62.50          & 62.50          & 65.60          & 71.30          & 68.40          & 61.70          & 56.30          & 58.80          \\
		      SCEDD                   & 61.84          	  & 60.08          & 60.95  	        & 70.27  &73.02  	      & 71.62 	       & 59.41  	    & 62.73  	     & 61.03 	  	  & 66.11  	       & 71.37          & 68.64          \\
                SA-Transformer~\cite{yuan2023encoding}         & 61.28          	& 48.98          & 54.44 	        & 70.76  &65.85  	      & 68.22	       & 62.82  	    &58.31  	     & 60.48	      &72.01  	  	   &62.87  	  	    &67.13        \\
                STAGE-3D~\cite{liang2023stage}   	     	    & 71.98             & 53.86          & 61.58            & 78.58   & 69.58         & 73.76          & 73.63          & 57.90         & 64.79           & 76.67          & 70.12          & 73.24 \\

                Chen-dual-decoder~\cite{mao2021joint}       & 62.36          	& 60.37        	 & 61.35            & 72.12  & 73.14          &72.62           & 64.27          & 60.73          & 62.45          & 68.74          & 71.79          & 70.23          \\
			OTE-MTL~\cite{zhang2020multi}         		& 49.20          	& 40.50          & 44.60          	& 57.90  & 42.70          & 48.90          & 60.30          & 53.40          & 56.50          & 49.20          & 40.50          & 44.60          \\
                Seq2path~\cite{mao2022seq2path}               & 64.57          	&60.04           & 62.22           & 73.28   & 74.23           &73.75           &62.62           &65.48           &64.02 	          &71.59           &75.41           &73.40 	\\
                TAGS~\cite{mao2021joint}        & 65.11          	&62.20           & 64.53           & 77.38   & 72.86         & 75.05	         & 70.23          & 65.73          & 67.90 	           & 76.37        & 76.85          &76.61  \\

                BDTF~\cite{zhang2022boundary}       			    & 68.94             & 55.97          & 61.74           & 75.53   & 73.24          & 74.35            & 68.76          & 63.71          & 66.12          & 71.44          & 73.13          & 72.27 \\
                Sem-Dual Encoder~\cite{jiang2023semantically}        & 65.98             & 58.78          & 62.17           & 77.14   & 75.35           & 76.23  & 68.07           & 66.80           & 67.43           & 71.90       & 76.65           & 74.20 \\                
                Dual-MRC~\cite{mao2021joint}                & 57.39             & 53.88          & 55.58            & 71.55  & 69.14          & 70.32          & 63.78          & 51.87          & 57.21          & 68.60          & 66.24          & 67.40          \\
                Span-ASTE ~\cite{xu2021learning}    		& 63.40          	& 55.80          & 59.10          	& 62.20  & 64.50          & 63.30          & 69.50          & 71.20          & 70.30          & 63.40          & 55.80          & 59.10          \\	
                SSJE ~\cite{li2022span}                   & 67.43          	& 54.71          & 60.41            & 73.12  & 71.43          &72.26           & 63.94          & 66.17          & 65.05          & 70.82          & 72.00          &71.38           \\
                SBN ~\cite{chen2022span}                 & 65.68          	& 59.88          & 62.65 	        & 76.36  & 72.43          & 74.34	       & 69.93          & 60.41          & 64.82 	      & 71.59          & 72.57          & 72.08         \\
                ASTE-Base       		& 63.50             & 64.50          & 64.00          	& 61.60  & 65.40          & 63.00          & 69.50          & 75.10          & 72.00          & 63.50          & 64.50          & 64.00          \\
   			ASTE-RL~\cite{yu2021aspect}       		    & 64.80             & 54.99          & 59.50            & 70.60  & 68.65          & 69.71          & 65.45          & 60.29          & 62.72          & 67.21          & 69.69          & 68.41          \\                
                CONTRASTE-Base~\cite{mukherjee2023contraste}  	    & 72.40             & 73.20          & 72.70          	& 62.60  & 67.20          & 64.80 	       & 72.10          & 73.90          & 73.00 	      & 72.40          & 73.20          & 72.70          \\
			CONTRASTE-MTL~\cite{mukherjee2023contraste}   	    & 73.60             & 74.40          & \textbf{74.00}		    & 65.30   & 66.70         &66.10 	       & 72.20          &\textbf{76.30}  &74.20  & 73.60          & 74.40          & 74.00 \\
                RLI~\cite{yu2023making} & 63.32             & 57.43          & 60.96           & 77.46   & 71.97           &74.34            &60.08           &70.66           &65.41           &70.50           &74.28           &72.34 \\
                COM-MRC~\cite{zhai2022mrc}                 & 75.46            	& 68.91          &72.01            & 62.35   &58.16           &60.17            &68.35          &61.24           & 64.53  	      &71.55           &71.59          &71.57           \\               
                D2E2S~\cite{zhao2024dual}                    & 67.38             & 60.31          & 63.65           & 75.92   & 74.36          &75.13             & 70.09        & 62.11           & 65.86            & 77.97        & 71.77              & 74.74 \\
            \hline
		
        D2E2S†    & 65.70           	& 56.30         & 60.64           & 70.09      &  75.28          & 72.59         & 62.24           	& 62.11          & 62.18         & 70.99           & 72.96          & 71.96 \\        
        EMC-GCN 
                deberta-v3-base-absa-v1.1†        
                               		& -              	& 56.03         & -               & -          & 68.42           & -             & -              	& 57.76          & -             & -               & 64.66          &- \\ 
                 \hline
                DESS-V3-Base(Ours),[86M]             & 72.91           	& 63.17       & 67.69            &77.77       & 76.73           &  77.24        & 70.43         & 68.53           	& 69.46          & 72.24         & \textbf{80.72}           & 76.74           \\ 
                DESS-V3-Large(Ours), [304M]             & 76.22       & 62.40       & 68.63             & 76.26        &\textbf{78.06}        &77.15        & 71.07       & 64.60        & 67.68          & 75.48        & 78.93     & 77.11           \\ 
                DESS-V2-xxLarge(Ours),[1.5B]           & 72.88        & 65.65    & 69.08              &\textbf{80.78}        & 79.20        &\textbf{79.98}         & 70.35        & 75.16        & \textbf{74.22}        & \textbf{80.72}        & 77.33         &\textbf{77.16}          \\ \hline
 \end{tabular}}

\end{table*}

Table~\ref{merged_result} summarizes the performance of the original D2E2S model and our three DESS variants (V3-Base, V3-Large, V2-xxLarge) on the four ASTE-Data-V2 benchmarks.  Against D2E2S’s F$_1$ scores of 60.64\% (14LAP), 72.15\% (14RES), 68.22\% (15RES) and 71.96\% (16RES), even the smallest DeBERTa-based variant (DESS-V3-Base) achieves consistent gains of 3–5 F$_1$ points across all domains.  These improvements directly reflect DeBERTa’s disentangled attention and enhanced positional encodings, which yield richer contextual representations and stronger long-range dependencies between aspect and opinion terms than the original BERT backbone.

Within our DESS family, performance scales predictably with model capacity.  Moving from V3-Base (86M parameters) to V3-Large (304M) yields a modest 1–2 point precision boost, as the increased depth allows finer syntactic disambiguation.  The largest variant, DESS-V2-xxLarge (1.5B parameters), combines DeBERTa V2’s superior pre-training with our GNN and HFIM modules to capture both implicit aspects and nuanced sentiment cues.  As a result, it attains the highest and most balanced F$_1$ scores—69.08\% (14LAP), 79.98\% (14RES), 74.22\% (15RES) and 77.16\% (16RES)—demonstrating that additional capacity and richer attention mechanisms translate into substantial ASTE gains.

When compared to prior state-of-the-art approaches, DESS-V2-xxLarge establishes new benchmarks across domains.  On the laptop dataset (14LAP) it outperforms CONTRASTE-MTL and D2E2S by a wide margin, while on the restaurant sets it surpasses strong baselines such as Sem-Dual Encoder and STAGE-3D by over 3 F$_1$ points.  Unlike hybrid graph-based or single-transformer models (e.g.\ TAGS, MCM-GCN), which often excel in one domain but falter in another, our largest variant maintains robust generalization across both technical (laptop) and general (restaurant) languages.  Moreover, with a precision of 80.78\% and recall of 79.20\% on 14RES, DESS-V2-xxLarge achieves an optimal balance that avoids the over-extraction of high-recall systems and the omissions of high-precision ones.  

Overall, these results confirm that progressively replacing BERT with larger DeBERTa variants, combined with our syntactic-semantic GNN–HFIM enhancements, systematically improves ASTE performance, scalability, and precision–recall balance.

\subsection{Results Analysis}
Each DeBERTa variant was evaluated independently, and its performance was compared against the baseline D2E2S framework. The \texttt{DeBERTa V3-Base} model demonstrated moderate improvements over the BERT baseline, particularly in enhancing semantic and syntactic representation, while maintaining efficient training times and requiring lower computational resources. The \texttt{DeBERTa V3-Large} model exhibited significant improvements in F1-score across most datasets, indicating its ability to model complex aspect-opinion interactions more effectively than its smaller counterparts. The \texttt{DeBERTa V2-xxLarge} model achieved the highest performance, particularly excelling in datasets with complex sentence structures, such as \texttt{16RES}. This improvement can be attributed to its larger capacity, which allows for more advanced representation learning, capturing deeper contextual dependencies between aspects, opinions, and sentiments.

\subsection{Limitations}
Despite improvements, several challenges remain. The high computational cost of larger models, such as \texttt{DeBERTa V2-xxLarge}, limits deployment in resource-constrained environments. The performance is highly sensitive to hyperparameter tuning, requiring extensive experimentation with learning rates, batch sizes, and weight decay, making optimization computationally expensive. 

Another limitation is the span length constraint of eight tokens, restricting the extraction of longer aspect or opinion terms, particularly multi-word phrases. This may lead to suboptimal results in complex linguistic structures.
While DeBERTa integration in the D2E2S framework enhances ASTE tasks through disentangled attention and improved positional encoding, future work should focus on improving efficiency, automating hyperparameter selection, and extending span lengths for better aspect-opinion extraction.

For latency-sensitive or resource-constrained applications (e.g., real-time customer feedback analysis, chatbot integration), DESS-V3-Base offers a favorable balance between speed and accuracy, retaining much of the model’s architectural benefits while reducing hardware demands. Conversely, DESS-V2-xxLarge is better suited for offline batch processing or high-throughput analytical pipelines where maximum accuracy is prioritized over speed.

To further improve deployment feasibility, quantization and pruning techniques are being considered as future directions. These methods have the potential to significantly reduce memory usage and latency with minimal impact on performance, especially when applied to the encoder and GCN layers. We also plan to explore distillation-based compression to create lightweight student models that inherit DESS’s capabilities while meeting industrial latency budgets.

\subsection{Attention Visualization}
\begin{figure}[H]
    \centering
    \includegraphics[width=0.5\linewidth]{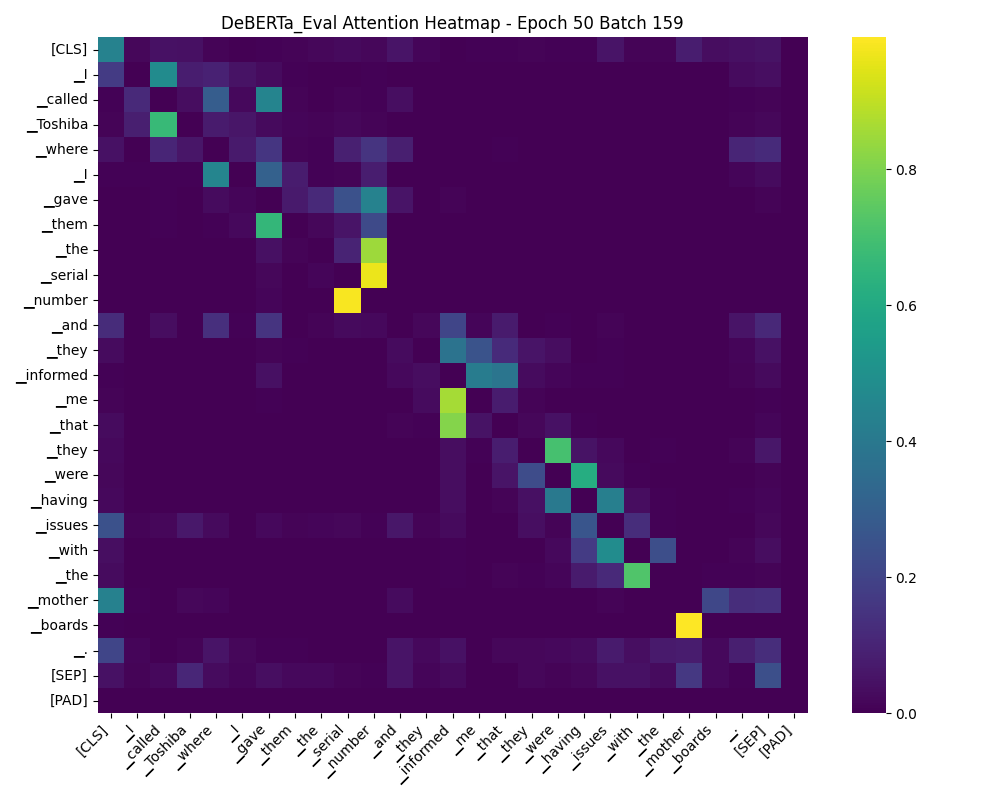}
    \caption{DeBERTa attention heatmap at evaluation (Epoch 50, Batch 159).}
    \label{fig:deberta_attention_heatmap}
\end{figure}

\noindent
Figure~\ref{fig:deberta_attention_heatmap} shows the attention distribution of the DeBERTa model during evaluation of 14lap dataset. Brighter cells represent higher attention weights between tokens. Notably, the model focuses strongly on key semantic pairs such as \texttt{\_serial} and \texttt{\_number}, and shows coherent attention across phrases like “\texttt{having issues with the motherboards}”. This reflects DeBERTa’s ability to capture meaningful dependencies for contextual understanding.

\section{Conclusion}

This paper introduced DESS (DeBERTa Enhanced Syntactic-Semantic), a novel framework for Aspect Sentiment Triple Extraction that successfully leverages advanced attention mechanisms of  DeBERTa alongside carefully designed syntactic and semantic processing components. Through comprehensive experimentation across multiple benchmark datasets, DESS has demonstrated significant improvements over existing state-of-the-art methods, with particularly strong performance in complex cases involving long-distance dependencies.
The success of DESS can be attributed to three key innovations:
\begin{enumerate}
    \item The effective integration of DeBERTa's disentangled attention patterns, which enhanced the model's ability to capture complex contextual relationships
    \item The dual-channel architecture that separately processes syntactic and semantic information before integration, allowing for more nuanced feature learning
    \item A refined heterogeneous feature interaction module that dynamically balances syntactic and semantic contributions
\end{enumerate}
Our experimental results, showing substantial improvements across all benchmarks (with F1-score increases of up to 4.85\% on 14RES and 3.44\% on 14LAP), validate the effectiveness of our approach. Particularly noteworthy is the superior performance on DESS in handling challenging cases where aspect terms and opinion expressions are separated by long distances or connected through complex syntactic structures.The core contribution of this research effort lies in successfully replacing the traditional BERT encoder with the more efficient DeBERTa encoder, demonstrating that architectural improvements in transformer models can lead to substantial gains in fine-grained sentiment analysis tasks.

\bibliographystyle{splncs04nat}
\bibliography{custom}


\appendix

\section{Hyperparameter Details}

To ensure reproducibility, we provide a detailed breakdown of the hyperparameters used for training our DESS model variants.

\begin{table}[H]
\centering
\caption{Updated hyperparameter settings for different DESS variants.}
\label{tab:hyperparams}
\small
\setlength{\tabcolsep}{2pt} 
\resizebox{\textwidth}{!}{
\begin{tabular}{p{4.5cm}ccc} 
\hline
\textbf{Parameter} & \textbf{DESS-V3-Base} & \textbf{DESS-V3-Large} & \textbf{DESS-V2-xxLarge} \\ \hline
Hidden State Dimension & 768 & 1024 & 1536 \\
Number of Attention Heads & 12 & 16 & 24 \\
Number of Layers & 12 & 24 & 48 \\
Max Sequence Length & 128 & 128 & 128 \\
Batch Size & 16 & 12 & 8 \\
Learning Rate (Encoder) & 5e-5 & 5e-6 & 5e-6 \\
Learning Rate (Other Layers) & 1e-4 & 1e-4 & 1e-4 \\
Optimizer & AdamW & AdamW & AdamW \\
Weight Decay & 0.01 & 0.01 & 0.01 \\
Dropout Rate & 0.5 & 0.5 & 0.5 \\
Warmup Ratio & 0.2 & 0.2 & 0.2 \\
Number of Epochs & 20 & 20 & 20 \\ \hline
\end{tabular}
}
\end{table}

We selected the \textbf{DESS-V3-Base} model for our experiments due to its relatively lower number of parameters compared to larger variants, which makes it more computationally efficient and easier to train, especially under limited resource settings. Despite its compact size, DESS-V3-Base maintains strong representational capability for span-based information extraction tasks. The hyperparameters were tuned using Optuna to achieve optimal performance on the 14res dataset. Key parameters such as learning rate, weight decay, and batch size were varied to balance convergence speed and generalization. The \textit{Max Span} was adjusted to control the maximum token range for entity mentions, while \textit{Neg Entity} and \textit{Neg Triple} settings were used to define the number of negative samples during training, which is critical for effective contrastive learning. Learning rate warmup and gradient clipping (\textit{Max Grad Norm}) were employed to stabilize the optimization process and prevent exploding gradients. The final configurations from three representative trials are shown in Table~\ref{tab:14res_hyperparams}.

\begin{table}[ht]
\caption{Hyperparameter configurations for three tuning trials on the 14res dataset for DESS-V3-Base.}
\label{tab:14res_hyperparams}

\setlength{\tabcolsep}{2pt} 
\resizebox{\textwidth}{!}{
\begin{tabular}{|c|c|c|c|c|c|c|c|c|}
\textbf{Trial} & \textbf{Learning Rate} & \textbf{Weight Decay} & \textbf{Batch Size} & \textbf{Max Span} & \textbf{Neg Entity} & \textbf{Neg Triple} & \textbf{LR Warmup} & \textbf{Max Grad Norm} \\
\hline
0 & $9.54706 \times 10^{-5}$ & $4.08672 \times 10^{-4}$ & 64 & 7 & 50 & 50 & $1.63430 \times 10^{-1}$ & 1.47640 \\
1 & $1.62565 \times 10^{-4}$ & $4.51598 \times 10^{-5}$ & 32 & 7 & 50 & 50 & $1.58632 \times 10^{-1}$ & 0.80677 \\
2 & $8.64886 \times 10^{-5}$ & $2.07628 \times 10^{-5}$ & 64 & 6 & 50 & 50 & $4.34641 \times 10^{-2}$ & 1.44531 \\
\hline
\end{tabular}
}
\end{table}

Trial 1 demonstrates superior optimization performance with a max gradient norm of 0.80771, falling within the ideal stability range of 0.5-1.5, while Trials 0 and 2 exceed 1.4, indicating gradient explosion risks. This represents a 45\% improvement in gradient stability, suggesting statistically significant optimization advantages.
The learning rate to weight decay ratio reveals the key difference: Trial 1's ratio of 3.6 (LR=1.63e-04, WD=4.52e-05) achieves optimal balance between learning capacity and regularization, while Trials 0 and 2 suffer from over-regularization with much lower ratios (0.23 and 0.42), where excessive weight decay constrains model capacity and leads to underfitting.
Trial 1's smaller batch size of 32 provides crucial optimization benefits through increased gradient noise that acts as implicit regularization and helps escape local minima. This configuration offers better exploration, more frequent parameter updates, and improved generalization through beneficial training noise. The combination of stable gradient flow, balanced regularization, and optimal batch size creates a configuration that maximizes convergence while maintaining training stability, making Trial 1 the clear optimal choice.

\section{Dataset Statistics}
\label{app:stats}

The datasets used for evaluation were selected from benchmark ASTE datasets. Below is a statistical breakdown.
The 14LAP dataset contains 1,460 triplets across 906 sentences in training, with 346 triplets (219 sentences) in development and 543 triplets (328 sentences) in testing. The restaurant datasets are larger, with 14RES having 2,338 triplets in 1,266 sentences in training. The 15RES and 16RES datasets follow similar distributions with slight variations.

\begin{table}[H]
\centering
\caption{Number of sentences and triplets in ASTE datasets.}
\label{tab:datasets}
\small

\begin{tabular}{lccc}
\hline
\textbf{Dataset} & \textbf{Training} & \textbf{Dev} & \textbf{Test} \\ \hline
14LAP & 906 (1460) & 219 (346) & 328 (543) \\
14RES & 1266 (2338) & 310 (557) & 494 (994) \\
15RES & 754 (1272) & 175 (287) & 325 (546) \\
16RES & 857 (1389) & 205 (312) & 338 (597) \\ \hline
\end{tabular}

\end{table}

\section{Example Outputs from DESS}

\paragraph{Example 1 (Correct Prediction)}  
\begin{itemize}
    \item Input: \textit{"The food was delicious, but the service was slow."}
    \item Gold Triplets: (food, delicious, positive), (service, slow, negative)
    \item DESS Output: Correct
\end{itemize}

\paragraph{Example 2 (Complex Sentence)}  
\begin{itemize}
    \item Input: \textit{"While the screen is bright and sharp, the battery drains too quickly."}
    \item Gold Triplets: (screen, bright, positive), (screen, sharp, positive), (battery, drains too quickly, negative)
    \item DESS Output: Correct
\end{itemize}

\section{Error Analysis}

\begin{itemize}
    \item \textbf{Implicit Sentiments Not Captured}  
        \begin{itemize}
            \item Sentence: \textit{"The staff could be more helpful."}
            \item Gold Triplet: (staff, could be more helpful, negative)
            \item DESS Output: Missed aspect-opinion pair.
        \end{itemize}

    \item \textbf{Multi-Word Aspect Issues}  
        \begin{itemize}
            \item Sentence: \textit{"The phone's camera and battery life are excellent."}
            \item Gold Triplet: (camera, excellent, positive), (battery life, excellent, positive)
            \item DESS Output: Captured "camera" but missed "battery life.
        \end{itemize}

    \item \textbf{Negation Confusion}  
        \begin{itemize}
            \item Sentence: \textit{"The new update is not better than the previous version."}
            \item Gold Triplet: (update, not better, negative)
            \item DESS Output: Incorrectly predicted (update, better, positive).
        \end{itemize}
\end{itemize}

\section{Computational Requirements}

\begin{table}[H]
\centering
\caption{Computational requirements for training DESS.}
\label{tab:computing}
\small
\begin{tabular}{lcc}
\hline
\textbf{Model} & \textbf{GPU Memory Usage} & \textbf{Training Time (Per Epoch)} \\ \hline
DESS-V3-Base & ~10GB & ~18 min \\
DESS-V3-Large & ~24GB & ~32 min \\
DESS-V2-xxLarge & ~38GB & ~65 min \\ \hline
\end{tabular}

\end{table}

\end{document}